\pdfoutput=1
\documentclass[man,floatsintext]{apa7}

\usepackage{amsmath,amssymb}
\usepackage{graphicx}
\usepackage{booktabs}
\usepackage{multirow}
\usepackage{adjustbox}
\usepackage{float}
\usepackage[natbibapa]{apacite}
\usepackage{xurl}
\hypersetup{hidelinks}
\graphicspath{{./images/}}

\title{From Text to Parameters: Predicting Item Parameters from Embedding Regularization with Reliability and Design Ceilings}
\shorttitle{From Text to Parameters}
\authorsnames[1,1]{Shi-Ting Chen, Jinsong Chen}
\authorsaffiliations{{Faculty of Education, The University of Hong Kong}}
\authornote{Shi-Ting Chen and Jinsong Chen contributed equally to this work and share first authorship. Corresponding author: Jinsong Chen (jinsong.chen@live.com).}

\abstract{Newly developed items must ordinarily be field tested before their psychometric properties are known, creating a cold-start problem for item calibration. Predicting item parameters from item features is a long-standing measurement problem dating back to the linear logistic test model (LLTM); modern text embeddings automate the design matrix that the LLTM tradition specified by hand. We propose an evaluation framework that combines regularized regression on item-text embeddings, repeated cross-validated $R^2$ reported with its resampling standard deviation, and two upper bounds on attainable performance: a reliability ceiling derived from the standard errors of the estimated parameters, and a design ceiling derived from a simulation-based power calibration of the prediction pipeline. Applying the framework to a mathematics item bank (EEDI) and a medical-licensure difficulty benchmark (BEA 2024), we find that item difficulty is substantially predictable from text (repeated cross-validated $R^2 = 0.53$, about 57\% of its reliability ceiling), whereas discrimination and pseudo-guessing appear progressively less predictable. Once read against the ceilings, however, this hierarchy is revealed to be largely a hierarchy of target reliability rather than of text signal: text recovers a nearly uniform 57--63\% of the reliable variance in every difficulty target, while the 3PL pseudo-guessing parameter has a reliability ceiling of essentially zero and is simply an unusable target at this calibration precision. On BEA, embedding-based regression matches leaderboard-level RMSE while explaining almost no variance, illustrating why scale-free metrics and explicit ceilings are needed to benchmark this task. All results use repeated cross-validation on the pooled items; we show that a single train--test split can inflate apparent accuracy by 0.1--0.15 in $R^2$, and discuss implications for calibration-support applications and for how item-difficulty benchmarks should be constructed and reported.}

\keywords{item difficulty modeling, text embeddings, regularized regression, linear logistic test model, item calibration, cross-validation}

\begin{document}
\maketitle

\section{Introduction}

Operational assessment programs rely on the continuous development and replenishment of item pools. However, newly developed items face a cold-start problem. Traditionally, estimating item parameters requires administering new items to a sample of test takers, collecting their responses, and then fitting a psychometric model, such as an item response theory (IRT) model, to obtain parameter estimates \citep{mccarthy2021jump}. This field testing process is costly, time-consuming, and may expose items before operational use \citep{yancey2024bert}. The burden is most acute exactly where the demand for fresh items is greatest: high-stakes programs that must retire exposed items quickly, and computerized adaptive testing (CAT) systems whose large item banks cannot function until every item carries calibrated parameters. Accurate item parameter prediction from item content would allow new items to enter operational use sooner, reduce pretest sample requirements, and support item development itself, for example by flagging items whose predicted difficulty departs from the intended blueprint. The measurement community has begun to take this prospect seriously: \citet{ulitzsch2026using} recently showed that even moderately accurate item parameter predictions, used as informative priors, can substantially reduce calibration sample requirements for a high-stakes admission test.

Predicting item properties from item features is not a new idea in psychometrics. The linear logistic test model \citep[LLTM;][]{fischer1973linear} decomposes item difficulty into a weighted sum of hand-specified cognitive operations, and the explanatory item response modeling tradition \citep{deboeck2004explanatory} generalizes this idea to arbitrary item covariates. Cognitive design system approaches use such feature--difficulty relationships to generate items with predictable parameters \citep{embretson1998cognitive}, and collateral information about items has long been used to sharpen or replace small-sample calibration \citep{mislevy1988exploiting,mislevy1993how}. The chief obstacle in this tradition has always been the design matrix: specifying, by hand, item features that capture what makes an item difficult. Modern text embeddings address exactly this obstacle. An embedding model maps the full item text to a high-dimensional vector that encodes semantic content, required knowledge, and surface form without manual feature engineering. Text-based item parameter prediction can therefore be understood as an automated, high-dimensional successor to the LLTM: the embedding supplies the design matrix, and a regression model supplies the feature weights \citep{alkhuzaey2024text}.

This reframing clarifies what the field needs next, because the recent machine-learning literature on item difficulty modeling faces several recurring challenges. First, embedding models and prediction architectures evolve quickly; results tied to one specific model risk obsolescence, so the durable contribution must be an evaluation framework that transfers to any present or future representation. Second, reported performance is difficult to interpret: systems can post competitive root mean squared error (RMSE) while explaining almost no variance in the target, as the BEA 2024 shared task illustrated, where the best system (RMSE = 0.299) barely improved on a mean-only baseline (RMSE = 0.311) \citep{yaneva2024findings,li2025item}. Third, most studies predict a single classical difficulty index treated as ground truth, even though operational calibration requires multiple IRT parameters (difficulty, discrimination, pseudo-guessing) and even though the targets themselves are estimates with quantifiable uncertainty. A near-zero prediction $R^2$ for a noisily estimated parameter may say more about the target than about the text.

The present study addresses these challenges with an evaluation framework built from four transparent components: (a) regularized regression \citep{hoerl1970ridge,tibshirani1996regression,zou2005regularization} on item-text embeddings, which handles the high-dimensional, highly correlated design matrix and composes readily with other ingredients such as generated rationales, hand-crafted features, or observation weights; (b) cross-validated $R^2$, reported with its variability over repeated resampling, as the primary scale-free criterion, together with the exact relationship linking $R^2$, RMSE, and the target standard deviation; (c) a \emph{reliability ceiling}, the largest population $R^2$ any predictor can attain against an estimated target, derived from the standard errors of the calibration; and (d) a \emph{design ceiling}, the largest cross-validated $R^2$ the finite design (given items, embedding matrix, estimator, and fold structure) can be expected to report even for a noise-free signal, obtained by a simulation-based power calibration. The two ceilings turn otherwise ambiguous results into interpretable ones: an observed $R^2$ near a ceiling indicates that text has extracted most of what is extractable, whereas a low observed $R^2$ beneath a high ceiling is genuine evidence of weak text signal.

Using item response data from the EEDI mathematics dataset \citep{wang2020instructions} and difficulty labels from the BEA 2024 shared task \citep{yaneva2024findings}, we evaluate prediction of classical test theory (CTT) difficulty and of 1PL, 2PL, and 3PL item parameters. This study addresses four research questions. First, to what extent can item text embeddings predict item difficulty under CTT and 1PL frameworks, how stable are these estimates across resampling, and how do they compare with the BEA 2024 difficulty benchmark once metrics are placed on a common footing? Second, does predictability differ across IRT parameters (difficulty, discrimination, and pseudo-guessing), and how much of the difference is attributable to the reliability of the targets rather than to the text? Third, do rationale-augmented item representations and additional NLP features add predictive information beyond the original item text? Fourth, does incorporating item parameter estimation uncertainty through observation weights improve prediction?

\section{Background and Related Work}

\subsection{Obtaining item parameters for newly developed items}
Newly developed assessment items usually require empirical response data before they can be used in IRT-based scoring, equating, or adaptive testing. Under traditional calibration, items are administered to examinees and item parameters are estimated from response data. This procedure provides statistically grounded estimates of item functioning, but it is costly, time-consuming, and can expose items before operational use.

Expert judgment is a common low-cost alternative during item development. Experienced teachers or content specialists review items and assign difficulty ratings based on their professional judgment \citep{ling2008expert}. However, expert judgments are subjective and do not consistently reproduce empirically estimated item statistics. \citet{bejar1983subject} examined subject-matter experts' assessment of item statistics and showed that expert judgments should not be treated as substitutes for empirical item analysis. \citet{attali2014estimating} further argued that comparative judgments may improve expert estimation of item difficulty, implying that conventional absolute difficulty ratings are limited and sensitive to elicitation format.

\subsection{From item features to item parameters: the psychometric lineage}
The idea that item parameters can be predicted from item properties has a fifty-year history in measurement. In the LLTM, \citet{fischer1973linear} modeled Rasch item difficulty as a linear combination of the cognitive operations an item requires, so that the difficulty of a new item can be computed from its design features before any response data are collected. The explanatory item response modeling framework of \citet{deboeck2004explanatory} places this idea in a generalized linear mixed modeling context in which any item covariate (content category, linguistic property, cognitive demand) may enter the item side of the model. \citet{embretson1998cognitive} showed that when difficulty is well predicted from theory-based item features, the feature--parameter mapping can drive item generation itself, producing items whose parameters are known by design. In parallel, a collateral-information tradition demonstrated that auxiliary item information can be exploited to estimate or update item parameters when response data are scarce: \citet{mislevy1988exploiting} used item features as priors for Rasch difficulties, and \citet{mislevy1993how} showed how tests can be linked with little or no response data by exploiting such side information. Most recently, \citet{ulitzsch2026using} operationalized this logic for modern item difficulty modeling, using text-based predictions to construct informative priors that reduce 3PL calibration sample requirements in a high-stakes application.

Two lessons from this lineage frame the present study. First, the binding constraint has always been the design matrix: the LLTM works only as well as the hand-specified features, and constructing them requires deep cognitive analysis that does not scale across item pools. Text embeddings remove this constraint by supplying a generic, automatically computed feature space. Second, the lineage makes clear that the practical payoff of prediction is not to replace calibration but to reduce its cost through priors, screening tools, or design feedback, so the appropriate evaluation standard is not perfect recovery but well-characterized, honestly reported predictive strength. Both lessons motivate the framework developed in Section 3.

\subsection{Machine learning and NLP approaches}
Text-based item difficulty modeling attempts to predict item difficulty or item parameters from item content before new response data are available. From a machine-learning perspective, a calibrated item pool can be treated as a labeled dataset, where item text provides the predictors, and estimated item statistics or IRT parameters provide the labels. A systematic review \citep{alkhuzaey2024text} identified four common steps in automatic difficulty prediction: ground truth labeling, preprocessing, feature extraction, and prediction.

Earlier studies relied heavily on manually engineered linguistic and surface features. \citet{beinborn2014predicting,beinborn2015candidate} predicted language-test item difficulty using linguistic features and machine-learning models, showing that item text contains useful information and that prediction results can reach the level of human experts. \citet{ha-etal-2019-predicting} predicted the difficulty of multiple-choice questions in a high-stakes medical examination using lexical, syntactic, and semantic features extracted from item text. \citet{benedetto2020r2de} proposed R2DE, an NLP-based framework for estimating IRT parameters of newly generated questions, including both difficulty and discrimination. Static word embeddings such as word2vec and GloVe \citep{mikolov2013efficient,mikolov2013distributed} were used to compute semantic similarity among item components \citep{hsu2018automated}, and contextualized encoders extended this to context-dependent representations. \citet{yancey2024bert} developed BERT-IRT, integrating BERT embeddings \citep{devlin2019bert} and engineered NLP features into an explanatory IRT framework to accelerate item piloting, and \citet{mccarthy2021jump} used BERT-based item features in a multi-task generalized linear model to jump-start item difficulty estimates for adaptive language testing.

More recent work uses large language models (LLMs) to enrich item representations or to simulate examinees. \citet{feng2025reasoning} proposed a reasoning- and sampling-augmented method for multiple-choice difficulty prediction, generating reasoning steps for each option and modeling option-selection likelihoods, which outperformed linear regression and fine-tuned Longformer baselines on the EEDI dataset. \citet{li2025item} fine-tuned small and large language models with data-augmentation strategies for difficulty-label imbalance; their best ensemble achieved RMSE = 0.2926 on the BEA 2024 benchmark, improving over the shared-task best of 0.2990. The BEA 2024 shared task itself \citep{yaneva2024findings} ranked systems by RMSE for predicting the difficulty of clinical multiple-choice questions; the best system achieved RMSE = 0.299, only slightly better than the DummyRegressor baseline of 0.311, and the overview concluded that differences among top systems were minor, an observation whose implications for evaluation metrics we return to throughout.

\subsection{Regularized regression for high-dimensional embeddings}
Modern text embeddings produce high-dimensional and often highly correlated predictors. In item-level datasets, the number of calibrated items is usually much smaller than the number of embedding dimensions. Ordinary least squares regression is therefore prone to overfitting and unstable coefficient estimates. Regularized regression provides a transparent and computationally efficient approach for this setting. Lasso regression applies an $L_1$ penalty and can perform variable selection by shrinking some coefficients exactly to zero \citep{tibshirani1996regression}. Ridge regression applies an $L_2$ penalty and is useful when many correlated predictors jointly contribute to prediction \citep{hoerl1970ridge}. Elastic Net combines both penalties \citep{zou2005regularization}. These methods can be efficiently implemented in the R package glmnet \citep{friedman2010regularization}.

Beyond computational convenience, regularized regression has two properties that suit it as the reference estimator in an evaluation framework. First, it is model-agnostic with respect to the representation: any embedding model, present or future, produces a matrix to which the same pipeline applies, so that representations can be compared on equal terms and the framework does not expire with any particular model. Second, it composes naturally with other ingredients of the prediction problem (augmented text, hand-crafted features appended to the design matrix, and observation weights reflecting target uncertainty), each of which is examined below.

For multi-parameter prediction, a further question is whether item parameters should be predicted separately or jointly. Multi-response Gaussian regression can predict multiple targets simultaneously using a shared predictor structure and group penalty \citep{yuan2006model}. This may be beneficial if difficulty, discrimination, and pseudo-guessing share common textual predictors; if the parameters relate to different aspects of item functioning, separate prediction may be less restrictive. The present study compares both strategies.

\section{An Evaluation Framework with Two Ceilings}

This section assembles the framework used in the empirical study: the prediction model (Section 3.1), the evaluation metrics and their exact interrelation (Section 3.2), and two upper bounds against which observed performance should be read (Sections 3.3 and 3.4).

\subsection{Embeddings as an automated design matrix}
Let $x_i \in \mathbb{R}^p$ denote the embedding of item $i$'s text and $y_i$ the item parameter target. In LLTM terms, $x_i$ plays the role of the row of the design matrix that \citet{fischer1973linear} specified by hand, with $p$ on the order of thousands rather than a handful of theorized operations. The coefficients are estimated by the Elastic Net objective
\begin{equation}
\hat{\beta} = \arg\min_{\beta_0,\beta} \left\{ \frac{1}{2n}\sum_{i=1}^{n} w_i (y_i - \beta_0 - x_i^T\beta)^2 + \lambda \left( \alpha \|\beta\|_1 + \frac{1-\alpha}{2}\|\beta\|_2^2 \right) \right\},
\label{eq:elnet}
\end{equation}
where $\lambda$ controls the strength of regularization, $\alpha$ the balance between the $L_1$ and $L_2$ penalties (Lasso: $\alpha=1$; Ridge: $\alpha=0$), and $w_i$ are optional observation weights (all 1 unless stated otherwise). The $L_1$ penalty performs implicit variable selection; the $L_2$ penalty distributes coefficient weight across correlated dimensions, which is often advantageous for embeddings. The regularization parameter $\lambda$ was selected by cross-validation within the training data using cv.glmnet \citep{friedman2010regularization,james2013introduction}; for Elastic Net, $\alpha$ was also selected by cross-validation over $\{0, 0.1, \ldots, 1\}$, so Elastic Net can select the Lasso or Ridge solution when the data favor a fully $L_1$- or $L_2$-penalized model.

\subsection{Evaluation metrics and the RMSE--SD--$R^2$ relationship}
Prediction performance was evaluated using RMSE and the coefficient of determination ($R^2$), computed on held-out data. Let $y_i$ denote observed values and $\hat{y}_i$ predictions for $i = 1, \ldots, n$, with $\bar{y}$ the mean of the observed values:
\begin{equation}\label{rmse}
RMSE = \sqrt{\frac{1}{n}\sum_{i=1}^{n} (y_i - \hat{y}_i)^2},
\qquad
R^2 =  1 - \frac{\sum_{i=1}^{n} (y_i - \hat{y}_i)^2}{\sum_{i=1}^{n} (y_i - \bar{y})^2}.
\end{equation}
$R^2$ compares the model against a mean-only baseline; negative values indicate predictions worse than the mean. With the standard deviation $SD$ of the observed values defined with denominator $n-1$, the three quantities are linked exactly by
\begin{equation}\label{relation}
R^2 =  1 - \frac{n}{n-1} \left( \frac{\mathrm{RMSE}}{\mathrm{SD}} \right)^2.
\end{equation}
Equation \ref{relation} clarifies two points that recur in this literature. First, RMSE is scale-dependent: two targets can have very different RMSE values but identical $R^2$ if their RMSE/SD ratios agree, so RMSE comparisons are meaningful only within a target scale. Second, an RMSE close to the target SD means the predictions are no better than the mean, which is the situation of the BEA 2024 leaderboard, where RMSE = 0.299 against a target SD near 0.31 implies $R^2 \approx 0.07$ at best. Reporting $R^2$ (or the RMSE/SD ratio) alongside RMSE prevents leaderboard-competitive error values from being mistaken for explanatory power.

Because a single train--test split yields a single draw from a distribution of possible results, we evaluate every model with repeated $K$-fold cross-validation on the pooled item set and report the mean and standard deviation of the out-of-fold RMSE and $R^2$ across repeated fold assignments. On item banks of a few hundred items, split-to-split variability is large enough to change substantive conclusions: for the best pipeline below, holdout $R^2$ for 1PL difficulty ranges from 0.39 to 0.57 across 20 random 70:30 splits of the same size (SD = 0.056). A single reported split can therefore fall well above or below the typical value, which is why we avoid single-split evaluation entirely and use it, if at all, only as a cautionary reference.

\subsection{The reliability ceiling: how well can a noisy target be predicted?}
The item parameters serving as targets are estimates, not true values. Write the calibration output for item $i$ as
\begin{equation}
\hat Y_i = T_i + e_i, \qquad e_i \sim (0,\ \mathrm{SE}_i^2), \qquad e \perp T,
\end{equation}
where $T_i$ is the true parameter and $\mathrm{SE}_i$ the standard error from the IRT fit. The variance of the estimates then contains the error variance, $\mathrm{Var}(\hat Y) = \mathrm{Var}(T) + \overline{\mathrm{SE}^2}$ with $\overline{\mathrm{SE}^2} = n^{-1}\sum_i \mathrm{SE}_i^2$. Even a perfect predictor of $T$ cannot explain the measurement noise $e$, so the largest population $R^2$ attainable against $\hat Y$ is the reliability of $\hat Y$:
\begin{equation}\label{eq:relceil}
R^2_{\mathrm{rel}}
  \;=\; 1 - \frac{\overline{\mathrm{SE}^2}}{\mathrm{Var}(\hat Y)}
  \;=\; \frac{\mathrm{Var}(T)}{\mathrm{Var}(T)+\overline{\mathrm{SE}^2}}.
\end{equation}
This ceiling depends on how precisely each item is calibrated (essentially, respondents per item), not on the number of items or predictors. It converts an ambiguous observation (``parameter $X$ is poorly predicted from text'') into a decidable one: if $R^2_{\mathrm{rel}}$ is itself low, the target is too noisy for \emph{any} predictor to do well, and the poor result reflects the calibration rather than the text. Prediction performance for a target should therefore be reported as a fraction of its reliability ceiling whenever standard errors are available.

\subsection{The design ceiling: simulation-based power calibration of cross-validated $R^2$}
A second, independent bound comes from the design: with $n$ items and $p \gg n$ embedding dimensions, even a noise-free linear signal cannot be recovered perfectly by cross-validated regularized regression, and cross-validated $R^2$ is biased downward at small $n$. We therefore calibrate the pipeline by parametric simulation, requiring only the embedding matrix $X$:
\begin{enumerate}
\item Column-standardize $X$ and compute the leading $m$ principal component scores $Z_{1:m}$, where $m$ is the assumed dimensionality of the signal. Placing the signal in the top-variance directions is the friendliest case for the estimator, so results are optimistic bounds.
\item Draw coefficients $\beta \sim N(0, I_m)$, form the standardized signal $s$ from $Z_{1:m}\beta$, and construct the synthetic target $y = s + \varepsilon$ with $\varepsilon_i \sim N(0, \sigma^2)$ and $\sigma = \sqrt{(1-R^2_{\mathrm{true}})/R^2_{\mathrm{true}}}$, so the population $R^2$ of the synthetic signal equals a chosen $R^2_{\mathrm{true}}$.
\item Run the identical estimator and $K$-fold cross-validation on $(X, y)$ and record the recovered cross-validated $R^2$; repeat over independent draws and average.
\end{enumerate}
The resulting map $\overline{R^2_{\mathrm{cv}}}(R^2_{\mathrm{true}}, m)$ answers: ``what would this design report if the truth were $R^2_{\mathrm{true}}$?'' Setting $R^2_{\mathrm{true}} = 1$ gives the outright \emph{design ceiling}: the best cross-validated $R^2$ the design can be expected to yield for an $m$-dimensional linear signal. Because the true $m$ is unknown, we report a grid (e.g., $m \in \{1, 5, 20\}$) rather than a single value, and we read observed results against the resulting band. The two ceilings compound: if text explains a fraction $\rho^\star \le R^2_{\mathrm{rel}}$ of the estimated target, the expected report is the power-attenuated $\overline{R^2_{\mathrm{cv}}}(\rho^\star)$. An observed value near the band means the design, not the text, is binding; a low observed value beneath a high band is genuine evidence of weak text signal. The calibration also shows directly why a null result on a small bank is uninformative, and how many items are needed before a given signal strength becomes detectable.

\section{Data and Method}

\subsection{Data sources}

\textbf{EEDI.} The first dataset was obtained from the NeurIPS 2020 Education Challenge \citep{wang2020instructions}. The dataset contains 948 multiple-choice questions (MCQs) originating from a mathematical education platform, Eedi (\url{https://eedi.com}). After careful review, we identified 361 text-only items (i.e., items that did not contain figures or diagrams) with response data from 6148 examinees. The original items were provided in image format. We therefore converted the item images into text using DeepSeek and manually checked the converted text against the original item images. The response data were used to estimate item parameters under CTT and IRT frameworks. After initial 2PL calibration, six items were removed because they showed poor or aberrant item functioning: three had negative discrimination estimates and three had extremely large difficulty estimates. The final analytic sample included 355 text-only items.

\textbf{BEA.} The second dataset was obtained from the BEA 2024 Shared Task \citep{yaneva2024findings}. The data consist of 667 MCQs sourced from the United States Medical Licensing Examination (USMLE), with CTT-based difficulty values provided for each item, partitioned into a training set ($n = 466$) and a test set ($n = 201$). The provided difficulty values are the linearly transformed proportion of correct responses across all examinees who answered the items, with higher values denoting more difficult items. Because examinee-level response data were not available, we did not estimate IRT parameters for BEA; the dataset was used for CTT-based difficulty prediction and, because the shared task ranked systems by RMSE, primarily as an external benchmark for metric comparison rather than as a strict test of cross-domain generalizability.

\subsection{Item parameter estimation and prediction targets}

For the EEDI dataset, item-level prediction targets were obtained under both CTT \citep{hambleton1993comparison, devellis2006classical} and IRT frameworks. The CTT difficulty target was defined as
\begin{equation}
d_j^{CTT}=1-p_j,
\end{equation}
where $p_j$ is the proportion of examinees who answered item $j$ correctly, so that higher values indicate more difficult items.

IRT item parameters were estimated using the R package mirt \citep{chalmers2012mirt}. The 1PL \citep{rasch1960probabilistic}, 2PL \citep{birnbaum1968some}, and 3PL \citep{birnbaum1968some, lord2012applications} models were fitted to the EEDI response matrix. Missing responses were left as missing and handled directly by mirt during estimation. All models converged. The models can be represented by the 3PL item response function
\begin{equation}
P(X_{ij}=1 | \theta_i) = c_j + (1 - c_j)\frac{\exp\bigl(a_j(\theta_i - b_j)\bigr)}{1 + \exp\bigl(a_j(\theta_i - b_j)\bigr)},
\end{equation}
where $X_{ij}$ is the response of examinee $i$ to item $j$, $\theta_i$ is examinee ability, $b_j$ is item difficulty, $a_j$ is item discrimination, and $c_j$ is the pseudo-guessing parameter. For the 1PL model, discrimination is constrained equal across items and $c_j=0$; for the 2PL model, $a_j$ and $b_j$ are estimated and $c_j=0$; for the 3PL model, all three are estimated.

The prediction targets were item difficulty $b_j$ (1PL, 2PL, 3PL), log-transformed discrimination $a_j^*=\ln(a_j)$ (2PL, 3PL), and logit-transformed pseudo-guessing $c_j^*=\ln\{c_j/(1 - c_j)\}$ (3PL). The transformations map positive and bounded parameters onto the unbounded scale assumed by the regression. Standard errors of the 1PL and 2PL item parameter estimates were obtained from mirt using SE = TRUE; standard errors for log-transformed discrimination were obtained by the delta method, $SE(\ln a_j) = SE(a_j)/a_j$, and for logit-transformed pseudo-guessing by $SE(\mathrm{logit}\,c_j) = SE(c_j)/\{c_j(1-c_j)\}$. For the 3PL model, the original calibration that produced the prediction targets was run without standard errors; a refit with SE = TRUE returned per-item standard errors for all 355 items via the Oakes method. This refit reached its EM iteration limit before full convergence, but it reproduced the original target estimates almost exactly (correlations: $b$ = 1.00, $a$ = .99, $c$ = 1.00), so the associated ceilings are treated as close approximations. The classical-difficulty reliability ceiling was obtained from per-item response counts $N_j$ in the response matrix via $SE_j^2 = p_j(1-p_j)/N_j$. These standard errors enter the reliability ceilings (Equation \ref{eq:relceil}) and the weighted regression analysis.

Because CTT difficulty is a monotone function of the total-score information that is sufficient for ability under the Rasch model, CTT and 1PL difficulty are expected to carry nearly identical ordering information when all examinees see comparable item sets; empirically, the two targets correlate at $r = 0.990$ (Spearman $\rho = 0.992$) in EEDI. We therefore treat results for the two targets as one finding rather than two.

\subsection{Item text representation}
For each item, the primary text input included the question stem, the correct answer, and all distractors, concatenated in a fixed template so that each embedding model received equivalent item information. The exact input templates and the rationale-generation prompt are given in Appendix \ref{app:templates}.

Five embedding models were evaluated: Qwen3-Embedding-8B \citep{zhang2025qwen3}, SFR-Embedding-2\_R \citep{meng2024sfr}, NV-Embed-v2 \citep{lee2025nv}, longformer-base-4096 \citep{beltagy2020longformer} and bert-large-uncased \citep{devlin2019bert}. These models differ in model size, maximum input length, and output dimensionality (Table \ref{tab:embedding_model}). The maximum token length for each model was determined automatically: if the model's configuration contained a sliding\_window attribute, that value was used; otherwise, the tokenizer's maximum length was used.

\begin{table}[!ht]
\centering
\caption{Specifications of selected embedding models\label{tab:embedding_model}}
\begin{adjustbox}{max width=\textwidth}\begin{tabular}{llcc}
\toprule
Model & Model size & Max sequence length (sliding window) & Output dimensionality \\
\midrule
Qwen3-Embedding-8B          & 8B      & 32768 & 4096 \\
SFR-Embedding-2\_R          & 7B       & 4096  & 4096 \\
NV-Embed-v2 & 8B & 4096 & 4096 \\
longformer-base-4096        & 148M       & 4096  & 768 \\
bert-large-uncased          & 336M    & 512    & 1024 \\
\bottomrule
\end{tabular}\end{adjustbox}
\end{table}

In addition to the original item-text representation, we evaluated two augmented representations. First, rationales were generated for each answer option following \citet{feng2025reasoning} using DeepSeek-V4-Pro: for the correct answer, the rationale explained the reasoning steps; for each distractor, why the option was incorrect. The rationale-augmented text concatenated the original item text with these option-level explanations. The bert-large-uncased model was excluded from this analysis due to its 512-token limit. Second, we appended hand-crafted linguistic features (number of sentences, nouns, unique nouns, prepositions, and Flesch--Kincaid readability) and mathematical features (number of numerical values, text-based numerical values, operators, and unique operators) to the rationale-augmented embeddings, to test whether traditional interpretable features add information beyond modern embeddings.

\subsection{Prediction strategies}
For 2PL and 3PL parameter prediction, we compared separate prediction, in which each parameter is modeled independently by Equation \ref{eq:elnet}, and joint prediction using multi-response Gaussian regression \citep{yuan2006model} as implemented in glmnet (family = mgaussian), which imposes a shared predictor structure across outcomes through a group penalty.

For weighted regression, three observation-weighting schemes based on the standard errors of the 1PL difficulty estimates were compared:
\begin{equation}
w_i^{(1)} = \frac{1}{SE_i}, \quad
w_i^{(2)} = \frac{1}{SE_i^2}, \quad
w_i^{(3)} = \frac{1}{SE_i^2 + \hat{\tau}^2},
\qquad
\hat{\tau}^2 = \max\left(0,\ \widehat{\mathrm{Var}}(b) - \frac{1}{n}\sum_{i=1}^{n} SE_i^2 \right).
\end{equation}
The third scheme is the standard variance-stabilized (random-effects) weight. Note that when $\hat{\tau}^2$ dominates $\overline{SE^2}$, as it does here (Section 5.1), $w^{(3)}$ is nearly constant across items and weighted regression is expected to approach the unweighted baseline; the comparison is therefore also a check on whether more aggressive weighting can be justified.

\subsection{Experimental design}
The analyses were organized into four stages, all conducted on the EEDI data unless noted.

\textbf{Evaluation design.} All models were evaluated by repeated $K$-fold cross-validation on the pooled item set ($K = 5$, 10 independent fold assignments), reporting the mean and standard deviation of the out-of-fold RMSE and $R^2$ across repetitions. For EEDI this pools all 355 items; for BEA it pools all 667 items. This design uses every item for both training and evaluation, avoids the large sampling variability of any single train--test partition (Section 3.2), and yields an explicit resampling standard deviation for each reported value. We deliberately do not report single-split results as findings: a companion inspection of the pre-specified split distributed with the original EEDI benchmark showed that it had been chosen, by a seed search, to maximize one pipeline's holdout accuracy, and it returns an $R^2$ (0.68 for 1PL difficulty) above the maximum of a 20-draw random-split reference distribution (mean 0.502, SD 0.056), precisely the inflation that repeated cross-validation removes.

\textbf{Stage 1: model and method selection; EEDI versus BEA.} The five embedding models and three regression methods were compared under repeated cross-validation for 1PL and CTT difficulty, establishing the best-performing combination. The same pipeline was applied to BEA. The selected combination (Qwen3-Embedding-8B with Ridge; Section 5.2) was then used exclusively in Stages 2--4, for simplicity and to avoid post hoc selection across the full grid at every stage.

\textbf{Stage 2: multi-parameter prediction.} 2PL targets (difficulty, log-discrimination) and 3PL targets (difficulty, log-discrimination, logit pseudo-guessing) were predicted under separate and joint strategies, and the resulting hierarchy of predictability was read against the reliability and design ceilings.

\textbf{Stage 3: augmentation.} Original item-text embeddings were compared with rationale-augmented embeddings, and with rationale-augmented embeddings plus hand-crafted NLP features, focusing on 1PL difficulty and 2PL difficulty and discrimination under separate prediction.

\textbf{Stage 4: weighted regression.} The three weighting schemes were compared for 1PL difficulty prediction, with both unweighted and weighted evaluation metrics:
\begin{equation}
\mathrm{RMSE}_{\text{weighted}} = \sqrt{\frac{\sum_{i} w_i (y_i - \hat{y}_i)^2}{\sum_{i} w_i}},
\qquad
R^2_{\text{weighted}} = 1 - \frac{\sum_{i} w_i (y_i - \hat{y}_i)^2}{\sum_{i} w_i (y_i - \bar{y}_w)^2},
\end{equation}
where $\bar{y}_w$ is the weighted mean of the observed values.

\section{Results}

\subsection{Descriptive statistics and ceilings}

\begin{figure}[!ht]
\centering
\includegraphics[width=0.6\textwidth]{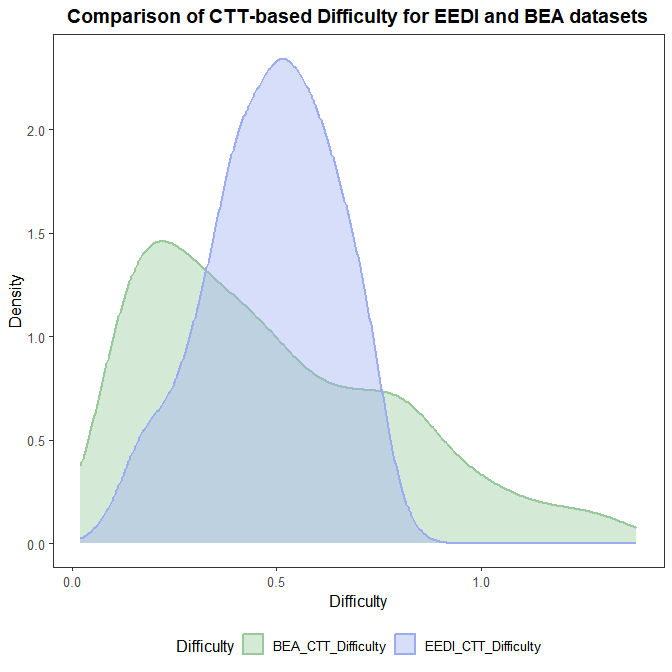}
\caption{Distribution of CTT-based difficulty values for EEDI and BEA}
\label{ctt_diff}
\end{figure}

\begin{figure}[!ht]
\centering
\includegraphics[width=0.6\textwidth]{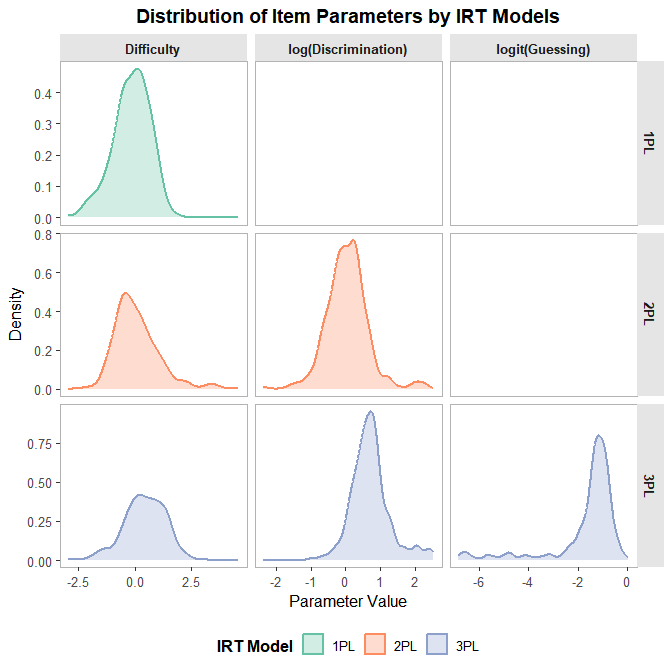}
\caption{Distribution of IRT parameters for EEDI}
\label{irt_diff}
\end{figure}

Table \ref{tab:summary_dataset} presents descriptive statistics for the two datasets together with the target-side reliability ceilings computed from Equation \ref{eq:relceil}. The density distributions of the CTT-based difficulty for EEDI and BEA are shown in Figure \ref{ctt_diff}, and the distributions of the IRT parameters for EEDI in Figure \ref{irt_diff}.

Three features of Table \ref{tab:summary_dataset} shape the interpretation of everything that follows. First, the reliability ceilings for CTT and 1PL difficulty are high (0.93 and 0.94): with an average of about 1600 responses per item, difficulty is precisely calibrated, so nearly all of its variance is in principle predictable and any shortfall must be attributed to the text side. Second, the ceiling falls to 0.76 for 2PL difficulty and to 0.57 for 2PL log-discrimination: almost half of the observed variance in log-discrimination is calibration noise, so even a perfect text-based predictor could not exceed $R^2 \approx 0.57$ against these estimates. Third, the between-item variance component $\hat\tau^2$ dominates the mean squared standard error for 1PL difficulty ($0.63$ vs.\ $0.04$), which is why the variance-stabilized weights $w^{(3)}$ are nearly uniform (Section 5.6).

\begin{table}[!ht]
\centering
\caption{Summary statistics and reliability ceilings for the prediction targets\label{tab:summary_dataset}}
\begin{adjustbox}{max width=\textwidth}\begin{tabular}{lccccc}
\toprule
Target & Dataset & SD & Mean $SE$ & $\overline{SE^2}$ & Reliability ceiling $R^2_{\mathrm{rel}}$ \\
\midrule
CTT difficulty & EEDI & 0.156 & 0.026\,$^a$ & 0.002 & 0.929 \\
CTT difficulty & BEA  & 0.309 & --\,$^b$ & --\,$^b$ & --\,$^b$ \\
$b$ (1PL) & EEDI & 0.816 & 0.128 & 0.040 & 0.940 \\
$b$ (2PL) & EEDI & 0.977 & 0.205 & 0.228 & 0.761 \\
$\ln(a)$ (2PL) & EEDI & 0.594 & 0.199 & 0.152 & 0.571 \\
$b$ (3PL)$^c$ & EEDI & 0.890 & 0.203 & 0.109 & 0.862 \\
$\ln(a)$ (3PL)$^c$ & EEDI & 0.562 & 0.276 & 0.165 & 0.430 \\
$\mathrm{logit}(c)$ (3PL)$^c$ & EEDI & 1.377 & 1.334 & 9.22 & $\approx 0$\,$^d$ \\
\bottomrule
\end{tabular}\end{adjustbox}
\begin{flushleft}
\footnotesize
Note. EEDI: $n = 355$ items (of 948; 361 text-only, 6 removed after 2PL screening), 6148 examinees. BEA: $n = 667$ items (466 train, 201 test). Reliability ceiling from Equation \ref{eq:relceil}; $SE(\ln a)$ by the delta method.
$^a$From per-item response counts, $SE_j^2 = p_j(1-p_j)/N_j$ (mean $N_j$ = 1641, range 7--3344).
$^b$Not computable: BEA provides only transformed difficulty labels without response counts.
$^c$From a 3PL refit with SE = TRUE (Oakes method); the refit reproduced the original target estimates at correlations of .99 or above, but its EM iteration limit was reached, so these ceilings are approximate. Delta method for $\ln(a)$ and $\mathrm{logit}(c)$.
$^d$The estimated reliability is negative ($-5.1$): the average sampling variance of $\mathrm{logit}(c)$ exceeds its between-item variance by a factor of about six, so the effective ceiling is zero.
\end{flushleft}
\end{table}

Table \ref{tab:design_ceiling} presents the design ceilings and the power-calibration map obtained by the simulation of Section 3.4 on the actual embedding matrices. Two implications follow. First, at $n = 355$ (EEDI) and $n = 667$ (BEA), the design ceilings are close to 1 even for a 20-dimensional signal, so both designs are adequately powered: near-zero observed $R^2$ on BEA cannot be excused as a small-sample artifact. Second, the power map shows only mild attenuation at these sample sizes: an injected true $R^2$ of 0.6 is recovered as 0.56 on the EEDI design, so the observed difficulty $R^2$ of 0.53 corresponds to a true linear signal strength of just under 0.6, clearly below the 0.94 reliability ceiling. The 0.6 row of the grid thus brackets the strongest result observed in this study and serves as its calibration reference. The binding constraint on EEDI difficulty prediction is therefore neither the design nor the calibration but the text representation itself, indicating genuine headroom for richer or fine-tuned representations and nonlinear prediction maps.

\begin{table}[!ht]
\centering
\caption{Design ceilings and power calibration: recovered cross-validated $R^2$ for synthetic signals of known strength embedded in the actual embedding matrices (Ridge, $K = 5$)\label{tab:design_ceiling}}
\begin{adjustbox}{max width=\textwidth}\begin{tabular}{lcccccc}
\toprule
& \multicolumn{3}{c}{EEDI (Qwen3-8B, $n=355$)} & \multicolumn{3}{c}{BEA (NV-Embed-v2, $n=667$)} \\
\cline{2-4}\cline{5-7}
$R^2_{\mathrm{true}}$ & $m=1$ & $m=5$ & $m=20$ & $m=1$ & $m=5$ & $m=20$ \\
\midrule
1.0 (ceiling) & 0.998 & 0.997 & 0.990 & 0.994 & 0.993 & 0.984 \\
0.8 & 0.775 & 0.772 & 0.747 & 0.768 & 0.767 & 0.757 \\
0.6 & 0.563 & 0.558 & 0.521 & 0.553 & 0.556 & 0.541 \\
0.4 & 0.357 & 0.351 & 0.308 & 0.346 & 0.353 & 0.338 \\
\bottomrule
\end{tabular}\end{adjustbox}
\begin{flushleft}
\footnotesize
Note. Each cell is the mean recovered cross-validated $R^2$ (Ridge, $K=5$) over 10 Monte Carlo replications of a synthetic target with the stated true population $R^2$, injected into the leading $m$ principal components of the actual embedding matrix. $m$ = assumed signal dimensionality. Monte Carlo SDs are 0.01--0.04 (largest at $R^2_{\mathrm{true}}=0.4$); full values in the online supplement. Entries are optimistic bounds (linear signal in top-variance directions).
\end{flushleft}
\end{table}

\subsection{Stage 1: Embedding model and method selection}

Table \ref{tab:eedi_first} presents the repeated cross-validation selection grid for 1PL and CTT difficulty in EEDI: five embedding models crossed with three regularized regression methods, each entry a mean over 10 fold assignments with its resampling standard deviation. Three patterns emerge. First, item difficulty is consistently predictable from item text embeddings across the whole grid. The best combination is Qwen3-Embedding-8B with Ridge regression ($R^2 = 0.533 \pm 0.038$ for 1PL, $0.528 \pm 0.037$ for CTT). Second, Ridge performs best or is matched by Elastic Net (whose cross-validated $\alpha$ frequently selects the Ridge solution), and both beat Lasso for the strongest models, consistent with a signal distributed across many correlated embedding dimensions rather than concentrated in a sparse subset. Third, the three large models (Qwen3, SFR, NV-Embed) clearly outperform the two small ones (BERT, Longformer). The near-identical $R^2$ for 1PL and CTT difficulty, despite RMSE values an order of magnitude apart, follows from Equation \ref{relation} and from the near-perfect correlation of the two targets ($r = 0.990$); we treat them as a single finding. Qwen3-Embedding-8B with Ridge is best or tied-best throughout and is used exclusively hereafter.

\begin{table}[!ht]
\centering
\caption{Stage 1 selection grid: repeated 5-fold cross-validation (10 fold assignments) for 1PL and CTT difficulty on the pooled 355 EEDI items; each cell is mean (SD) across repetitions\label{tab:eedi_first}}
\begin{adjustbox}{max width=\textwidth}\begin{tabular}{llcccc}
\toprule
\multirow{2}{*}{Method}  & \multirow{2}{*}{Model} & \multicolumn{2}{c}{1PL Difficulty} & \multicolumn{2}{c}{CTT Difficulty} \\
\cline{3-6}
& & RMSE & $R^2$ & RMSE & $R^2$\\
\midrule
\multirow{5}{*}{Lasso}
& Qwen3-Embedding-8B & 0.571 (0.025) & 0.508 (0.045) & 0.111 (0.004) & 0.496 (0.040)\\
& SFR-Embedding-2\_R & 0.609 (0.018) & 0.441 (0.034) & 0.118 (0.004) & 0.432 (0.041)\\
& NV-Embed-v2 & 0.616 (0.023) & 0.427 (0.043) & 0.121 (0.004) & 0.401 (0.042)\\
& Longformer-base-4096 & 0.654 (0.028) & 0.355 (0.056) & 0.126 (0.004) & 0.346 (0.047)\\
& bert-large-uncased & 0.667 (0.014) & 0.329 (0.029) & 0.125 (0.003) & 0.355 (0.034)\\
\midrule
\multirow{5}{*}{Ridge}
& Qwen3-Embedding-8B & 0.556 (0.022) & 0.533 (0.038) & 0.107 (0.004) & 0.528 (0.037)\\
& SFR-Embedding-2\_R & 0.567 (0.020) & 0.516 (0.035) & 0.109 (0.004) & 0.512 (0.032)\\
& NV-Embed-v2 & 0.576 (0.018) & 0.499 (0.032) & 0.112 (0.003) & 0.488 (0.030)\\
& Longformer-base-4096 & 0.673 (0.012) & 0.318 (0.025) & 0.129 (0.002) & 0.315 (0.025)\\
& bert-large-uncased & 0.638 (0.010) & 0.386 (0.019) & 0.121 (0.002) & 0.395 (0.019)\\
\midrule
\multirow{5}{*}{Elastic Net}
& Qwen3-Embedding-8B & 0.566 (0.024) & 0.516 (0.042) & 0.109 (0.004) & 0.515 (0.037)\\
& SFR-Embedding-2\_R & 0.567 (0.020) & 0.515 (0.035) & 0.109 (0.004) & 0.511 (0.033)\\
& NV-Embed-v2 & 0.579 (0.019) & 0.495 (0.033) & 0.112 (0.004) & 0.484 (0.033)\\
& Longformer-base-4096 & 0.646 (0.025) & 0.371 (0.049) & 0.125 (0.005) & 0.358 (0.053)\\
& bert-large-uncased & 0.641 (0.014) & 0.380 (0.028) & 0.121 (0.003) & 0.395 (0.029)\\
\bottomrule
\end{tabular}\end{adjustbox}
\end{table}

\subsection{EEDI versus BEA: same pipeline, different explainable variance}

Table \ref{tab:eedi_bea} places the EEDI and BEA results on a common footing, expressing each in terms of RMSE, the RMSE/SD ratio, and $R^2$, alongside the published BEA reference points; all of our own values are repeated cross-validation means. The all-model BEA breakdown appears in Table \ref{tab:cvgrid}; performance varied little across embedding models ($R^2$ from 0.01 to 0.03).

Two conclusions follow. First, on BEA, embedding-based regularized regression is competitive with the leaderboard: our best cross-validated RMSE (0.304) sits between the shared-task winner (0.299) and the dummy baseline (0.311), and close to the best published result (0.2926 by fine-tuned ensembles; \citealp{li2025item}). Yet the RMSE/SD ratio is essentially 1 for every system including the winner: by Equation \ref{relation}, the implied $R^2$ is about 0.07 for the shared-task winner and about 0.11 for the best published fine-tuned ensemble. In other words, no known method, including heavy fine-tuning, explains appreciable variance in the BEA difficulty scale; our near-zero $R^2$ replicates the state of the art rather than falling short of it. Given the high design ceiling for BEA in Table \ref{tab:design_ceiling}, this is genuine evidence that the USMLE difficulty labels carry little text-recoverable signal, plausibly reflecting the homogeneous, highly able examinee population, range restriction in operational item pools, and clinical-reasoning (rather than surface-text) determinants of difficulty. Second, on EEDI the same pipeline attains an RMSE/SD ratio of 0.69 and repeated cross-validated $R^2$ of 0.53, demonstrating that when the item bank and target permit, the framework does recover substantial signal, making EEDI the more informative benchmark for method development.

\begin{table}[!ht]
\centering
\caption{EEDI versus BEA CTT-difficulty prediction on a common metric footing\label{tab:eedi_bea}}
\begin{adjustbox}{max width=\textwidth}\begin{tabular}{llccccc}
\toprule
Dataset & System / design & $n_{\mathrm{train}}$ / $n_{\mathrm{test}}$ & SD & RMSE & RMSE/SD & $R^2$ \\
\midrule
EEDI & Qwen3+Ridge, repeated CV & 355 (pooled) & 0.156 & 0.107 & 0.69 & 0.528 \\
BEA & Qwen3+Ridge, repeated CV & 667 (pooled) & 0.309 & 0.306 & 0.99 & 0.016 \\
BEA & bert-large+Ridge, repeated CV & 667 (pooled) & 0.309 & 0.304 & 0.98 & 0.031 \\
BEA & Shared-task winner \citep{yaneva2024findings} & 466 / 201 & 0.309 & 0.299 & 0.97 & --\,$^a$ \\
BEA & Dummy (mean) baseline \citep{yaneva2024findings} & 466 / 201 & 0.309 & 0.311 & 1.01 & $\approx 0$ \\
BEA & Fine-tuned ensemble \citep{li2025item} & 466 / 201 & 0.309 & 0.2926 & 0.95 & --\,$^a$ \\
\bottomrule
\end{tabular}\end{adjustbox}
\begin{flushleft}
\footnotesize
Note. Our own rows are repeated cross-validation means on the pooled items; leaderboard rows are as published on the official 466/201 split (test SD 0.311, essentially equal to the pooled SD). $^a$Not reported by the source; the implied $R^2$ from Equation \ref{relation} is shown in the text.
\end{flushleft}
\end{table}

Table \ref{tab:cvgrid} breaks the BEA result out by embedding model under the same repeated cross-validation protocol. No embedding explains appreciable variance ($R^2$ from 0.01 to 0.03, all standard deviations spanning zero), so the near-zero BEA result is not specific to any one representation. Read against the high BEA design ceiling in Table \ref{tab:design_ceiling}, this confirms that the EEDI--BEA gap is a property of the data (the USMLE difficulty labels simply carry little text-recoverable signal) rather than of the evaluation design.

\begin{table}[!ht]
\centering
\caption{Repeated 5-fold cross-validation (Ridge, 10 fold assignments) for BEA CTT-difficulty prediction, all five embedding models on the pooled 667 items; mean (SD) across repetitions\label{tab:cvgrid}}
\begin{adjustbox}{max width=\textwidth}\begin{tabular}{lcc}
\toprule
Embedding model & RMSE & $R^2$ \\
\midrule
Qwen3-Embedding-8B  & 0.306 (0.001) & 0.016 (0.008) \\
SFR-Embedding-2\_R  & 0.306 (0.002) & 0.017 (0.012) \\
NV-Embed-v2         & 0.307 (0.001) & 0.012 (0.008) \\
longformer-base-4096 & 0.304 (0.002) & 0.026 (0.011) \\
bert-large-uncased  & 0.304 (0.001) & 0.031 (0.008) \\
\bottomrule
\end{tabular}\end{adjustbox}
\begin{flushleft}
\footnotesize
Note. BEA CTT SD = 0.309. The corresponding EEDI all-model results appear in the CTT columns of Table \ref{tab:eedi_first}.
\end{flushleft}
\end{table}

\subsection{Stage 2: The predictability hierarchy across IRT parameters}

Table \ref{tab:hierarchy} is the central result of the study: for each prediction target, it reports the repeated cross-validation performance of the selected pipeline (Qwen3-Embedding-8B with Ridge, separate prediction), the reliability ceiling of the target, and the fraction of reliable variance recovered. The joint-versus-separate comparison for the 2PL and 3PL parameter sets appears in Appendix Table \ref{tab:jointsep}.

\begin{table}[!ht]
\centering
\caption{Predictability hierarchy: repeated 5-fold cross-validation (10 repetitions) with Qwen3-Embedding-8B and Ridge, read against target reliability\label{tab:hierarchy}}
\begin{adjustbox}{max width=\textwidth}\begin{tabular}{lcccccc}
\toprule
Target & RMSE mean (SD) & $R^2$ mean (SD) & $R^2_{\mathrm{rel}}$ & $R^2 / R^2_{\mathrm{rel}}$ \\
\midrule
CTT difficulty & 0.107 (0.004) & 0.528 (0.037) & 0.929 & 0.57 \\
$b$ (1PL) & 0.556 (0.022) & 0.533 (0.038) & 0.940 & 0.57 \\
$b$ (2PL) & 0.705 (0.024) & 0.478 (0.037) & 0.761 & 0.63 \\
$\ln(a)$ (2PL) & 0.500 (0.009) & 0.289 (0.026) & 0.571 & 0.51 \\
$b$ (3PL) & 0.630 (0.018) & 0.496 (0.030) & 0.862 & 0.58 \\
$\ln(a)$ (3PL) & 0.544 (0.005) & 0.060 (0.018) & 0.430 & 0.14 \\
$\mathrm{logit}(c)$ (3PL) & 1.306 (0.007) & 0.097 (0.009) & $\approx 0$ & --\,$^b$ \\
\bottomrule
\end{tabular}\end{adjustbox}
\begin{flushleft}
\footnotesize
Note. 3PL ceilings from the SE = TRUE refit (see Table \ref{tab:summary_dataset} notes). $^b$Undefined: the estimated ceiling is zero or negative, so no reliable variance is available to recover; the small positive observed $R^2$ is within what refit approximation error and the skewed SE distribution allow.
\end{flushleft}
\end{table}

Three findings emerge. First, the raw hierarchy is clear: difficulty is the most predictable target under every model ($R^2$ = 0.48--0.53 across 1PL, 2PL, and 3PL), 2PL log-discrimination is substantially less predictable (0.29), and the 3PL nuisance parameters are weakest (log-discrimination 0.06; logit pseudo-guessing 0.10). The pseudo-guessing $R^2$ is small but its resampling interval sits just above zero, so text carries a faint but non-null signal for the raw guessing estimates.

Second, and this is what the ceilings add, much of the hierarchy is an artifact of target reliability. Across all four difficulty targets the ceiling-adjusted recovery is remarkably uniform: text recovers 57\% (CTT), 57\% (1PL), 63\% (2PL), and 58\% (3PL) of the reliable variance in difficulty. For 2PL log-discrimination the ceiling is only 0.57, and text recovers 51\% of what is attainable. On the ceiling-adjusted metric, 2PL discrimination is nearly as text-recoverable as difficulty, and the dramatic raw gap (0.53 vs.\ 0.29) mostly reflects noisier calibration of $a$ rather than absence of textual signal. The 3PL nuisance parameters complete the picture. The ceiling for 3PL log-discrimination drops to 0.43 (of which text recovers only 14\%), and for logit pseudo-guessing the estimated ceiling is effectively \emph{zero}: the average sampling variance of $\mathrm{logit}(c)$ exceeds its between-item variance by a factor of about six, so the target contains essentially no reliable between-item signal for \emph{any} predictor to recover. Consistent with this, the 2PL and 3PL log-discrimination estimates correlate at only $r = 0.16$, reflecting the well-known $a$--$b$--$c$ trade-offs of 3PL calibration at this examinee sample size. The raw ordering ``difficulty $\gg$ discrimination $\gg$ guessing'' thus conflates two mechanisms, text signal and target noise, and the ceiling-adjusted column separates them: difficulty prediction is limited by the text representation, whereas the apparent unpredictability of the 3PL nuisance parameters is overwhelmingly a property of the calibration, not of the text.

Third, joint multi-response prediction offered no advantage over separate prediction and was harmful for difficulty (Appendix Table \ref{tab:jointsep}): under repeated cross-validation, joint 3PL difficulty $R^2$ falls to 0.369 (SD = 0.018) versus 0.496 (SD = 0.030) for separate prediction, while the weaker parameters are unchanged within resampling error. The group-penalty structure, which forces common predictor selection across targets, is too restrictive when the targets differ sharply in text recoverability, and it dilutes the difficulty signal without helping discrimination or guessing. Separate prediction is therefore the recommended default.

\subsection{Stage 3: Rationale and feature augmentation}

Because the effect is small, only the paired resampling design can resolve it. Augmenting the item text with LLM-generated option rationales produced a small but consistent improvement for 1PL difficulty: across the 10 matched fold assignments, mean $R^2$ rose to 0.557 (SD = 0.025) with rationales from 0.533 (SD = 0.038) without, with a correspondingly lower RMSE (0.542 vs.\ 0.556); the improvement held in the great majority of matched repetitions. Adding the nine hand-crafted linguistic and mathematical features on top of the rationale-augmented embeddings changed essentially nothing further ($R^2$ = 0.558, SD = 0.025). Substantively, LLM-generated rationales add a modest amount of difficulty-relevant information beyond the raw item text (on the order of $\Delta R^2 \approx 0.02$), whereas traditional NLP surface features add none, consistent with the original embedding already capturing the surface properties those features encode. An effect of this size (about two-thirds of the resampling SD) would be easily lost in single-split evaluation, underscoring the value of the paired repeated-CV design for detecting genuine but small gains.

\subsection{Stage 4: Weighted regularized regression}

Table \ref{tab:weight} reports the weighted-regression results for 1PL difficulty under repeated cross-validation. The variance-stabilized scheme $w^{(3)}$ performed on par with the unweighted baseline ($R^2$ = 0.533), while the more aggressive schemes $w^{(1)}$ and $w^{(2)}$ were clearly worse. This ordering is expected from Table \ref{tab:summary_dataset}: because $\hat\tau^2 \approx 0.63$ dominates $\overline{SE^2} \approx 0.04$, the weights $w^{(3)}$ are nearly constant, so the random-effects weighting correctly recognizes that calibration error is a minor component of target variance here; the aggressive inverse-variance schemes overweight a subset of precisely estimated items and harm generalization. Weighting would matter more in banks calibrated on small examinee samples, where $\overline{SE^2}$ rivals $\hat\tau^2$, which is exactly the setting quantified by the reliability ceiling.

\begin{table}[!ht]
\centering
\caption{Weighted regularized regression for 1PL difficulty prediction (EEDI, Qwen3-Embedding-8B), repeated 5-fold cross-validation (10 fold assignments); mean (SD) of $R^2$\label{tab:weight}}
\begin{adjustbox}{max width=\textwidth}\begin{tabular}{llcccc}\toprule
\multirow{2}{*}{Method} &\multirow{2}{*}{Weight}& \multicolumn{2}{c}{Unweighted Metrics} & \multicolumn{2}{c}{Weighted Metrics} \\
\cline{3-6}
 & & RMSE & $R^2$ & RMSE & $R^2$\\
\midrule
\multirow{3}{*}{Lasso}
&  $1 / SE$ & 0.587 & 0.481 (0.037) & 0.544 & 0.515 (0.042) \\
& $1 / SE^2$ & 0.627 & 0.408 (0.037) & 0.538 & 0.470 (0.051) \\
& $1 / (SE^2 + \hat{\tau}^2)$ & 0.568 & 0.513 (0.043) & 0.565 & 0.520 (0.045) \\
\midrule
\multirow{3}{*}{Ridge}
&  $1 / SE$ & 0.575 & 0.501 (0.035) & 0.523 & 0.551 (0.042) \\
& $1 / SE^2$ & 0.608 & 0.443 (0.032) & 0.501 & 0.542 (0.041) \\
& $1 / (SE^2 + \hat{\tau}^2)$ & 0.557 & 0.532 (0.038) & 0.553 & 0.541 (0.039) \\
\midrule
\multirow{3}{*}{Elastic Net}
&  $1 / SE$ & 0.576 & 0.499 (0.032) & 0.526 & 0.546 (0.038) \\
& $1 / SE^2$ & 0.609 & 0.441 (0.029) & 0.505 & 0.533 (0.038) \\
& $1 / (SE^2 + \hat{\tau}^2)$ & 0.564 & 0.519 (0.038) & 0.560 & 0.528 (0.039) \\
\bottomrule
\end{tabular}\end{adjustbox}
\end{table}

\section{Discussion and Conclusion}

This study reframed text-based item parameter prediction as the modern continuation of a long psychometric tradition, the LLTM with an automated, embedding-based design matrix, and proposed an evaluation framework combining regularized regression, repeated cross-validated $R^2$, and two explicit ceilings on attainable performance. Applying the framework to a mathematics item bank and a medical-licensure benchmark yields five conclusions.

First, item difficulty is substantially predictable from item text, but honest uncertainty quantification matters. Under repeated cross-validation, Qwen3-Embedding-8B with Ridge regression recovered a mean $R^2$ of 0.533 (SD = 0.038) for 1PL difficulty. The importance of the resampling design is underscored by what a single split can do: holdout $R^2$ across 20 random 70:30 splits ranged from 0.39 to 0.57, and a performance-based split selected for high accuracy can return 0.68, above the maximum of the random-split distribution. A spread this wide can reverse method comparisons and inflate absolute claims by 0.1--0.15 in $R^2$, so we report all results as repeated cross-validation with resampling SDs and use no single split as a finding. This point bears directly on how item-difficulty benchmarks should be released. A single fixed train--test partition is valuable for reproducibility and cross-study comparability (this is what makes BEA useful), but only if the partition is drawn independently of any model's performance. A split selected to maximize one pipeline's holdout accuracy is not method-neutral: it rewards systems whose representations resemble the one used for selection and yields optimistic estimates for all. When a fixed benchmark split is desired, it should be drawn by a method-agnostic rule (a documented random seed, or matching on the target difficulty to balance the train and test distributions), and paired with a recommended repeated cross-validation protocol on the pooled items so that absolute performance is reported without selection bias. A direct check confirms the logic: a split constructed to minimize the difficulty distance between train and test (Kolmogorov--Smirnov statistic 0.03, test SD 0.83 versus 0.82 overall) yields a single-split $R^2$ of 0.51 for 1PL difficulty, essentially the repeated cross-validation mean of 0.53, and far below the 0.68 of the performance-selected split. A distribution-matched split reduces the sampling variability of the estimate but does not, and should not, raise it; only selection on performance does that. Balancing rules that compress the test-set difficulty variance can even depress $R^2$ (a decile-stratified split gave 0.46), reaffirming that $R^2$ must be read jointly with the target SD (Equation \ref{relation}).

Second, the predictability hierarchy (difficulty $>$ discrimination $>$ pseudo-guessing) is real in raw terms but is largely a hierarchy of \emph{target reliability}, not of text signal. Reading observed $R^2$ against the reliability ceilings shows that text recovers a strikingly uniform 57--63\% of the reliable variance in every difficulty target (CTT, 1PL, 2PL, 3PL) and 51\% in 2PL discrimination; on the ceiling-adjusted metric, discrimination is nearly as text-recoverable as difficulty, and the dramatic raw gap (0.53 vs.\ 0.29) mostly reflects the noisier calibration of $a$. At the extreme, the 3PL pseudo-guessing parameter has an estimated reliability ceiling of essentially zero: its between-item variance is smaller than its average squared calibration error, so its near-zero prediction $R^2$ certifies an unusable target, not a failure of text. Benchmarks that treat estimated parameters as exact ground truth will systematically understate predictive validity for noisily calibrated targets; parameter-specific reliability ceilings should accompany parameter-specific performance, and targets whose ceilings approach zero should be excluded from prediction benchmarks altogether.

Third, the BEA analysis illustrates why RMSE-based leaderboards can mislead. Our simple pipeline matches leaderboard RMSE while explaining almost no variance, and so does everything else published on this benchmark, including heavily fine-tuned systems. Given the high design ceiling of the BEA embedding matrix, the conclusion is that the benchmark's difficulty labels carry little text-recoverable signal, not that any particular method failed. Scale-free metrics ($R^2$, RMSE/SD) and design-ceiling checks should be standard practice when constructing or interpreting such benchmarks.

Fourth, the augmentation and weighting extensions illustrate how the framework separates real effects from noise. Rationale augmentation yielded a small but consistent gain under repeated cross-validation ($\Delta R^2 \approx 0.02$ for 1PL difficulty), an effect about two-thirds the size of the resampling SD that a single-split evaluation would likely miss; hand-crafted NLP features added nothing beyond the embeddings; and SE-based observation weighting performed exactly as the variance decomposition predicts when calibration is precise: the principled random-effects weights are nearly uniform, and aggressive inverse-variance weighting harms generalization. The ceilings and resampling variability provide the yardstick that distinguishes genuine improvements from noise, a yardstick against which future extensions (fine-tuned representations, nonlinear prediction heads, multimodal encoders) can be judged.

Fifth, the practical value of the achieved accuracy should be understood through the calibration-support lens of the collateral-information tradition. \citet{ulitzsch2026using} demonstrate that predictions of the present accuracy class, used as informative priors, materially reduce calibration sample requirements; predicted difficulty can likewise support item screening, difficulty banding, and pretest design planning. Rather than aiming to replace empirical calibration, text-based prediction is best understood as reducing its cost, which is also the appropriate criterion for judging whether an $R^2$ in the range observed here is ``good enough.''

Several limitations should be acknowledged. First, the 3PL reliability ceilings rest on a refit whose EM iteration limit was reached before full convergence (though it reproduced the original target estimates at correlations of .99 or above and returned standard errors for every item); the 3PL ceilings should therefore be read as close approximations rather than exact values. Second, the EEDI analysis was limited to text-only items; results do not generalize to items in which figures or diagrams are central, and multimodal embeddings remain to be examined. Third, embeddings were used without task-specific fine-tuning, and only linear prediction maps were considered; the design-ceiling analysis suggests headroom that nonlinear or fine-tuned approaches might claim. Fourth, BEA provided only transformed difficulty labels, precluding IRT-parameter prediction and reliability analysis on that dataset. Finally, an interpretability extension is worth pursuing: replacing raw embedding dimensions with contextual similarity scores between item embeddings and keyword embeddings from an external domain corpus \citep{chen2025documents} would trade some predictive accuracy for named, inspectable predictors, potentially allowing Lasso-selected keywords to characterize \emph{what} makes items difficult; preliminary work in this direction is left to future study.

\section*{Data and Code Availability}
The complete analysis pipeline (R code for the regularized regression evaluation, the repeated cross-validation protocol, the reliability and design ceilings, and the Python code for building item documents and extracting embeddings), together with the fold-assignment seeds and derived result files sufficient to verify every reported table, is available for masked review through an anonymized OSF view-only link: \url{https://osf.io/yhgzw/overview?view_only=f9adc60686eb4fc98ed7ddcd0be9abf0}. The project will be made public and citable upon acceptance. Because the item texts, response data, and difficulty labels of both datasets are subject to their sources' terms, the original data are not redistributed. The EEDI items and response data are publicly available from the NeurIPS 2020 Education Challenge (\url{https://eedi.com/projects/neurips-education-challenge}; \citealp{wang2020instructions}), and the BEA 2024 shared task data can be requested from the task organizers (\url{https://sig-edu.org/sharedtask/2024}; \citealp{yaneva2024findings}). The package includes step-by-step instructions that reproduce the full pipeline, from raw data to every table, once those datasets are obtained.

\appendix

\section{Item text templates and rationale-generation prompt}\label{app:templates}

\subsection{Embedding input templates}
Each item was converted to a single document before embedding. For EEDI items, the document was constructed as follows (fields in braces are filled per item; blank lines separate the parts):

\begin{flushleft}
\ttfamily
Question: \{item stem\}\\[2pt]
Correct answer: \{correct answer\}\\[2pt]
Wrong answer 1: \{distractor 1\}\\[2pt]
Wrong answer 2: \{distractor 2\}\\[2pt]
Wrong answer 3: \{distractor 3\}
\end{flushleft}

For the rationale-augmented representation, the following blocks were appended to the same document:

\begin{flushleft}
\ttfamily
Rationale for correct answer: \{rationale\}\\[2pt]
Rationale for wrong answer 1: \{rationale\}\\[2pt]
Rationale for wrong answer 2: \{rationale\}\\[2pt]
Rationale for wrong answer 3: \{rationale\}
\end{flushleft}

For BEA items, the same structure was used with the item stem and the keyed answer text, followed by one ``Wrong answer $k$'' block for each non-empty option other than the key (USMLE items have up to ten options). The identical document was supplied to every embedding model, so that model comparisons reflect the representation rather than the input.

\subsection{Rationale-generation prompt}
Option-level rationales were generated one item at a time with DeepSeek-V4-Pro, following the reasoning-step design of \citet{feng2025reasoning}, using the following instruction:

\begin{flushleft}
\ttfamily
You are an experienced mathematics teacher. For the multiple-choice question below, first explain, step by step, how a student should reason to reach the correct answer. Then, for each wrong answer, explain the specific misconception or computational error that would lead a student to choose it. Keep each explanation concise (two to four sentences) and do not restate the question.\\[4pt]
Question: \{item stem\}\\[2pt]
Correct answer: \{correct answer\}\\[2pt]
Wrong answer 1: \{distractor 1\}\\[2pt]
Wrong answer 2: \{distractor 2\}\\[2pt]
Wrong answer 3: \{distractor 3\}
\end{flushleft}

The generated rationales were inserted into the rationale-augmented template above without further editing.

\section{Joint versus separate multi-parameter prediction}
Table \ref{tab:jointsep} reports the joint-versus-separate comparison for the selected pipeline (Qwen3-Embedding-8B with Ridge) under the same repeated cross-validation protocol used throughout, for the 2PL and 3PL parameter sets. Separate prediction models each parameter independently; joint prediction uses multi-response Gaussian regression (group penalty) across the parameter set.

\begin{table}[!ht]
\centering
\caption{Joint versus separate prediction of 2PL and 3PL parameters, Qwen3-Embedding-8B with Ridge, repeated 5-fold cross-validation (10 fold assignments); mean (SD) of $R^2$\label{tab:jointsep}}
\begin{adjustbox}{max width=\textwidth}\begin{tabular}{lllcc}
\toprule
Model & Parameter & Target & Separate $R^2$ & Joint $R^2$ \\
\midrule
2PL & difficulty & $b$ & 0.478 (0.037) & 0.472 (0.036) \\
2PL & discrimination & $\ln(a)$ & 0.289 (0.026) & 0.284 (0.035) \\
\midrule
3PL & difficulty & $b$ & 0.496 (0.030) & 0.369 (0.018) \\
3PL & discrimination & $\ln(a)$ & 0.060 (0.018) & 0.065 (0.026) \\
3PL & pseudo-guessing & $\mathrm{logit}(c)$ & 0.097 (0.009) & 0.086 (0.017) \\
\bottomrule
\end{tabular}\end{adjustbox}
\begin{flushleft}
\footnotesize
Note. Separate-prediction values match the corresponding rows of Table \ref{tab:hierarchy}. Joint prediction uses glmnet family = mgaussian.
\end{flushleft}
\end{table}

\bibliographystyle{apacite}
\bibliography{bibliography}

\end{document}